\documentclass{article}
\usepackage{spconf,amsmath,graphicx, threeparttable}
\usepackage{url}

\graphicspath{{Figures/}}

\begin{document}
\title{IvaNet: Learning to jointly detect and segment objets with the help of Local Top-Down Modules}
\name{Shihua Huang\thanks{The work is done during Shihua's undergraduation ($2014.9\sim2018.6$) in Northeastern University where he worked with Prof. Lu.}, Lu Wang}
\address{Northeastern University \\
Shenyang, Liaoning province, China \\
wanglu@ise.neu.edu.cn}
\maketitle

\begin{abstract}
Driven by Convolutional Neural Networks, object detection and semantic segmentation have gained significant improvements.
However, existing methods on the basis of a full top-down module have limited robustness in handling those two tasks simultaneously.
To this end, we present a joint multi-task framework, termed IvaNet.
Different from existing methods, our IvaNet backwards abstract semantic information from higher layers to augment lower layers using local top-down modules.
The comparisons against some counterparts on the PASCAL VOC and MS COCO datasets demonstrate the functionality of IvaNet.
\end{abstract}
\begin{keywords}
Object detection, Semantic segmentation, Local Top-Down module, Multi-task framework.
\end{keywords}

\section{Introduction}
\label{sec:intro}
Object detection and semantic segmentation play pivotal roles in image understanding that can be applied to numerous applications, e.g. automated driving.
Driven by the Convolutional Neural Networks (CNNs), those two tasks have gained significant improvement.
Due to the existence of pooling and stride-convolution layers in bottom-up network, the resolutions of higher layers is smaller.
As a result, those layers contain more abstract semantic information about objects since they have larger receptive field, but they will lose the spatial information that good for locating gradually.

\begin{figure}[!htbp]
\centering
    \includegraphics[width=\linewidth]{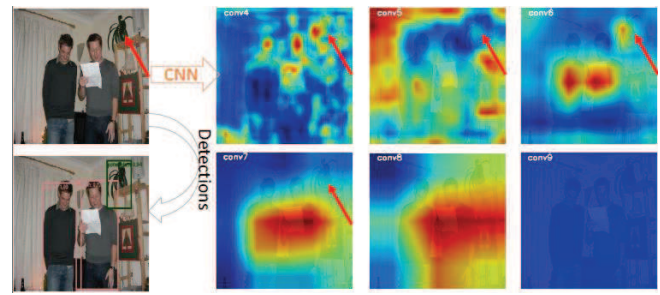}
    \vspace{-0.7cm}
   \caption{Selected visual feature maps from SSD300~\cite{ssd}.}
\label{fig:visual_features}
\vspace{-0.5cm}
\end{figure}

In order to better solve object detection and semantic segmentation tasks that require semantic information as well as spatial information, a number of full top-down modules are proposed to backward the semantic information from higher layers into lower layers where there is much more detailed information~\cite{dssd, deconvnet, blitznet}.
Despite that these methods have achieved encouraging results, they may introduce much useless information into some lower layers, which may degrade their performance.
It can be observed from Fig.~\ref{fig:visual_features} that the information of the plant after those two persons vanishes and is covered by larger objects (persons) gradually as the receptive field becomes larger, which means the information backward from some higher layers is meaningless for some lower layers that are used to detect and segment small objects.
To this end, a local top-down module (LTD)~\cite{Local_Top_Down} has been proposed and demonstrated its functionality for single shot object detector~\cite{ssd} recently.

This paper studies to apply the LTD module to a multi-task framework, called IvaNet, which is designed for effectively detecting and segmenting objects simultaneously.
To be specifical, we adopt a LTD module to integrate the information from two succeeding convolutional layers for each lower layer.
In this way, we can construct a local top-down network on the basis of a deep bottom-up neural network, ImageNet pre-trained ResNet50~\cite{resnet} is used in this study, and two task-specific heads are builded on the top of the top-down network.
We note that the idea behind our detector is the same with SSD~\cite{ssd} while the segmentation head is a simple FCN~\cite{fcn}.
Extensive experimental results on the PASCAL VOC and MS COCO datasets show that the local top-down achieves superior results on the task of semantic segmentation, which demonstrates its functionality.
Moreover the proposed multi-task learning improves the performance of each task.

\label{sec:framework}

\section{Related work}
\label{sec:related_work}
\subsection{Object Detection}
Object detection aims at locating objects with bounding boxes and classifying them into corresponding class.
The existing object detectors can be categorized into two branches, R-CNN based~\cite{rcnn, fast, faster} and SSD based~\cite{ssd, dssd, Local_Top_Down}.
R-CNN based methods use a third part method to pre-select anchors, for instance, R-CNN~\cite{rcnn} use the Selective Search~\cite{ss}.
Compared with R-CNNs, Single Shot Detector (SSD)~\cite{ssd} is fast and robust to multi-scale object detection.
There are some variants of SSD for improving the robustness by introducing the abstract semantic information from higher layers into lower ones via a top-down network.
Instead using a full top-down module that will introduce much useless information as DSSD~\cite{dssd}, LTD-SSD~\cite{Local_Top_Down} proposes a local top-down module and achieves better results.
Specifically, each prediction layer is integrated only with the upsampled features from its two succeeding layers.
\subsection{Semantic Segmentation}
The task of semantic segmentation is learning to classify each pixel of the input image into corresponding class.
FCN~\cite{fcn} is one of the pioneers that extended the convolutional model used for image-level classification to per-pixel classification by replacing all fully connected layers with convolutional layers.
Instead using a single bilinear interpolation layer to upsample the segmentation results to original size, DeconvNet~\cite{deconvnet} opts to use a deep learnable  deconvolution network.
Besides, DeepLab V2~\cite{deeplabv2} proposes a Atrous Spatial Pyramid Pooling module that can build a spatial pyramid without changing the resolutions of the feature maps.
While PSPNet\cite{pspnet} adopts a pyramid pooling module to aggregate contextual information from different regions.

\subsection{Multi-task learning}
A number of methods for multi-task learning have emerged after UberNet~\cite{ubernet}, which enables 7 computer vision tasks can be handled simultaneously with a single complex model.
Mask R-CNN~\cite{mask} augments the Faster R-CNN with a instance segmentation prediction branch and shows compelling results on object detection and instance segmentation.
Different from Mask R-CNN, the BlitzNet~\cite{blitznet} is a SSD-based multi-task framework for object detection and semantic segmentation that maintains a far superior speed.

\section{IvaNet}
\label{sec:framework}
\begin{figure*}[!htbp]
\centering
    \includegraphics[width=0.9\linewidth]{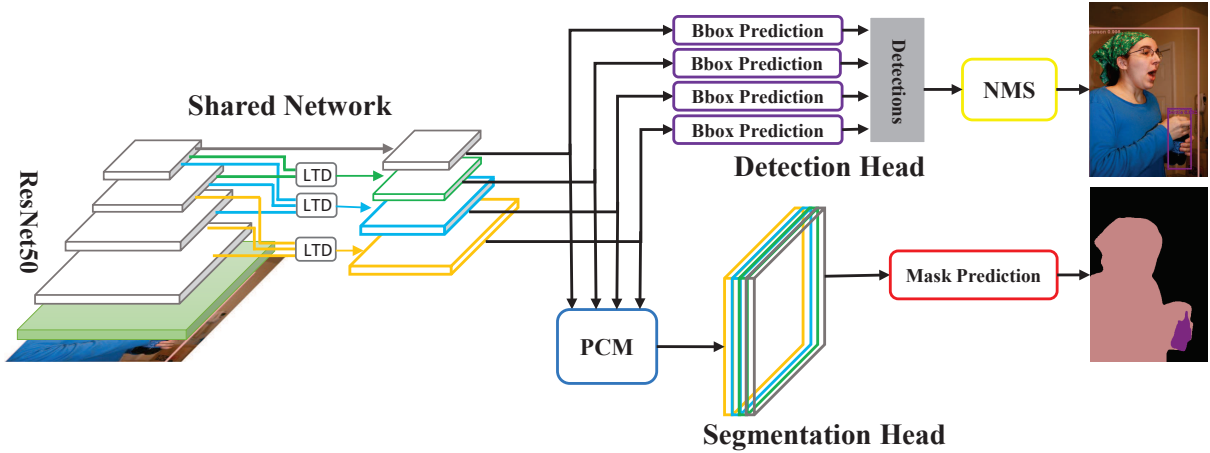}
    \vspace{-0.45cm}
   \caption{The overall structure of the proposed IvaNet, where the PCM represents a Pyramid Convolutional Module.}
\label{fig:whole_structure}
\vspace{-0.3cm}
\end{figure*}

\subsection{Architecture}
The overall architecture is shown in Fig.~\ref{fig:whole_structure}, IvaNet consists of a shared convolutional feature extraction network and two task-specific heads on top of it.

\textbf{Shared Network:} We use some ImageNet pre-trained layers from the ResNet50~\cite{resnet} to construct a base bottom-up network.
Specifically, the parameters after block4 all are dropped.
Furthermore, the same Local top-down module (LTD) presented in LTD-SSD~\cite{Local_Top_Down} is adopted to construct a top-down feature pyramid.
We note that the output channel of eah LTD is restricted to 384 by $1\times 1$ convolutional layer, which is different from LTD-SSD.

\textbf{Object Detection Head:} There are a number of Bounding Box Prediction (BBbox Prediction) branches with the identical network structures in Object Detection Head.
They are used to work at feature pyramid for multi-scale objects detection efficiently.
Each BBox Prediction contains a paired of detection-specific classification and location regression layer.
To be specific, they both are a convolutional layer whose kernel size is $3\times 3$ with output channel = $C\times A$ and $4\times A$ for classification and location regression, respectively, where $C$ is the number of object classes, e.g. it is 21 for PASCAL VOC dataset (one is background), and it used as a consistent denotation in the following unless otherwise specified, while $A$ denotes the number of default anchors in each cell, and $4$ represents two coordinates of each anchor.
Furthermore, we adopt the hard negative example mining to balance the ratio between positive and negative examples.
Finally, the Non-Maximum Suppression (NMS) is employed as the post-processing method to eliminate redundant detection results.

\textbf{Semantic Segmentation Head:} It can be observed that the semantic segmentation head includes a Pyramid Convolutional Module (PCM) and a mask prediction branch.
Instead using different pooling kernel sizes and strides to get multi-scale feature maps as Pyramid Pooling Module in PSPNet~\cite{pspnet}, we use the pre-computed feature pyramid from the local top-down network.
After that, they will be upsampled to pre-defined size and concatenated.
Before be used to predicted the semantic mask with the mask prediction branch, the concatenation will be input to a $3\times 3$ convolutional layer.
The mask prediction branch only has a convolutional layer with kernel size and output channel is set to $3 \times 3$ and $C$, respectively.

\subsection{Objective}
For detection, we use the same objective presented in SSD as follow,
\begin{eqnarray*} \label{eq:det_loss}
L_{det} =& \frac{1}{N}\left(L_{cls}(x, s) + L_{loc}(x,p,g)\right),
\end{eqnarray*}
where $L_{cls}$ is a classification loss that defined as the softmax loss over multiple classes scores $s$, and the localization loss $L_{loc}$ is a Smooth L1 loss between the predicted box $p$ and the groundtruth box $g$ parameters, and $N$ denotes the number of samples that used as the same meaning in the following formula~\ref{eq:seg_loss}.
While the segmentation loss is the cross-entropy between predicted and the target class distribution of pixels:
\begin{eqnarray*} \label{eq:seg_loss}
L_{seg}=& -\frac{1}{N}\sum_{\mathbf{X}}[y\times ln(y) + (1-y)\times ln(1-a)],
\end{eqnarray*}
where $\mathbf{X}$ is a set of samples, and $y$ is the target while $a$ is the predicted.
The final objective is defined as,
\begin{eqnarray*}\label{eq:objective}
L=& L_{det} + W \times L_{seg},
\end{eqnarray*}
where $W$ is a hyper-parameter that controls the relative importance of segmentation loss compared to the detection loss, and set to 1.
\subsection{Implementation Details}
We implement our model with TensorFlow~\cite{tensorflow} and it will be available at \url{https://github.com/Hshihua/IvaNet}.
The same data augmentation methods as SSD~\cite{ssd} is adopted, including random crop, horizontal flip, and so on.
For all experiments, the Adam optimizer~\cite{Adam} with $\beta_{1}=0.5, \beta_{2}=0.999$ is used to train our IvaNet, and the initial learning rate is set to 10$^{-4}$, which will be decreased twice by a factor 10.
Moreover, the mini-batch size is set to 32 or 16 when the resolution of the input image is 300 or 512, respectively.

\section{Experiments}
\label{sec:experiments}
In this section, we first introduce the datasets and metrics used in this paper simply.
Afterwards, the functionality of our proposed IvaNet is validated by the comparisons with some counterparts on various datasets.
Finally, there are some analyses about the possible reasons behind of our failures.
\subsection{Datasets and metrics}

\textbf{PASCAL VOC.}
The VOC2007 and VOC2012 are two of active datasets in detection and segmentation tasks.
Both datasets have thousands of images over 20 object classes.
Since there are only a set of images from VOC2012 are annotated with semantic masks in both datasets, which is less effective for semantic segmentation evaluation.
We augment them with extra annotations provided by~\cite{extra_ann}, denoted as VOC2012 train-aug.

\textbf{Microsoft COCO.}
The MS COCO dataset~\cite{mscoco} includes 80 categories of objects for object detection and instance segmentation.
There are hundreds of thousands of annotated images.
To get the semantic segmentation annotations from the given instance segmentation annotations, we use the tool provided by Nikita Dvornik et.al.~\cite{blitznet}.

The quality of predicted segmentation masks is measured with mean Intersection over Union (mIoU) in all datasets while we evaluate the detection results from PASCAL VOC or MS COCO with mean Average Precision (mAP) or AP$^{0.5:0.95}$ (AP, for simply), respectively.

\subsection{Results on the PASCAL VOC}
\label{sec:pascal_voc}
For all experiments on PASCAL VOC datasets, the max training iterations are set to 65k and 75k for models with input size as 300$\times$300 and 512$\times$512, respectively.
They decrease their initial learning rate at 35k and 45k steps, respectively, and the learning rates both are decreased anther time after 15k steps.

\textbf{Ablation Study.}
This part demonstrates the effectiveness of the LTD modules and validates a multi-task model is better than the single one. Three variants of the IvaNet that alate LTD modules, segmentation head and detection head, called IvaNet$_{NO}$, IvaNet$_{det}$ and IvaNet$_{seg}$, respectively.
From Tab.~\ref{tb:ablation}, we can see that the IvaNet without LTD modules degrades its performance significantly on both tasks.
Besides, the IvaNet improves the IvaNet$_{det}$ and IvaNet$_{seg}$ with the max improvements are 0.5\% and 1.7\%, respectively.
Furthermore, IvaNet only adds small extra time consumption to IvaNet$_{det}$ after building the segmentation head along with the detection head.
\begin{table}[htb]
\centering
\begin{tabular}{c|c|c|c}
\hline
Method & 300$\times$300 & 512$\times$512 &FPS (300$\backslash$512) \\
\hline
IvaNet$_{NO}$  &70.3/72.9 &77.1/75.2 &- \\
IvaNet$_{det}$ & 76.2/- &79.3/-  & 36.5$\backslash$25.5 \\
IvaNet$_{seg}$ &-/74.2 & -/76.2 & 44.6$\backslash$31.5\\
IvaNet  & 76.2/75.9 & 79.8/76.8 & 32.5$\backslash$24.5\\
\hline
\end{tabular}
\caption{Alation results on PASCAL VOC2012. The n and m from n/m are denoted as the mAP and the mIoU, respectively, and used the same in the following, while the runtime is tested using a single 1070TI GPU.}
\label{tb:ablation}
\vspace{-0.3cm}
\end{table}

\textbf{Comparison.}
As shown in Tab.~\ref{tb:pascal_voc} obviously, our proposed IvaNet has achieved superior or comparable results when compared to some single-task models.
Besides, we also compare the IvaNet with an existing multi-task model that exploits a full top-down module, BlitzNet.
Our IvaNet has nearly the same performance as BlitzNet~\cite{blitznet} on detection task, but IvaNet has achieved much better results on segmentation task and outperforms by 1.1\%.
\begin{table}[!htb]
\centering
\begin{threeparttable}
\begin{tabular}{c|c|c|c}
\hline
Method &Backbone &VOC2007\dag &VOC2012 \\
\hline
\tiny{detectors:} & & & \\
Faster RCNN~\cite{faster} &VGGNet16 & 73.2/-    &70.4/- \\
SSD300~\cite{ssd} &VGGNet   &77.2/- &75.8/- \\
DSSD321~\cite{dssd} &ResNet101  &78.6/- &76.3/- \\
LTD-SSD300~\cite{Local_Top_Down} &VGGNet    &79.4/- &76.7/- \\
SSD512~\cite{ssd} &VGGNet   &79.8/- &78.5/- \\
DSSD513~\cite{dssd} &ResNet101  &81.5/- &80.0/- \\
LTD-SSD512~\cite{Local_Top_Down} &VGGNet    &81.8/- &79.7/- \\
\hline
\tiny{segmentations:}   & & & \\
FCN~\cite{fcn}  &VGGNet     &-/-        &-/62.2 \\
DeconvNet~\cite{deconvnet}       &VGGNet     &-/-        &-/69.6 \\
Deeplab-v2~\cite{deeplabv2}      &VGGNet     &-/69.0     &-/- \\
GCN+BR~\cite{gcn}          &ResNet50   &-/72.3     &-/- \\
GCN+BR~\cite{gcn}          &ResNet101  &-/74.7     &-/- \\
\hline
\tiny{multi-task:} & & & \\
BlitzNet300~\cite{blitznet} &ResNet50   &78.7/75.3*   &76.7/75.7 \\
IvaNet300   &ResNet50   &78.5/75.5*   &76.2/75.9 \\
BlitzNet512~\cite{blitznet} &ResNet50   &81.5/75.7*   &79.7/76.7 \\
IvaNet512   &ResNet50   &81.4/76.9*   &79.8/76.8 \\
\hline
\end{tabular}
\begin{tablenotes}
\footnotesize
\item[\dag] The performance of detection and segmentation are tested on VOC2007 test set and VOC2012 val set, respectively.
\item[*] Models are trained with VOC07 trainval + VOC12 train-aug instead.
\end{tablenotes}
\end{threeparttable}
\caption{Comparison results on PASCAL VOC.}
\label{tb:pascal_voc}
\vspace{-0.5cm}
\end{table}

\subsection{Results on the MS COCO}
\label{sec:ms_coco}
All models are trained on the trainval35k for 700k iterations, and the learning rates are decreased at 400k and 500k steps.
Different from the PASCAL VOC, all trained images are both annotated with bounding boxes and semantic mask that is derived from the public instance mask.
It seems that the proposed IvaNet consistently outperforms the BlitzNet significantly on segmentation task from Tab.~\ref{tb:mscoco}, while keeps comparable results on detection task.
\begin{table}[!htb]
\centering
\begin{tabular}{c|c}
\hline
Method & minival (AP/mIoU)\\
\hline
BlitzNet300~\cite{blitznet} &29.7/52.8  \\
IvaNet300   & 29.7/55.0 \\
\hline
\end{tabular}
\caption{Comparison results on MS COCO. }
\label{tb:mscoco}
\vspace{-0.3cm}
\end{table}

\subsection{Visual Illustration}
\label{sec:visual_ill}

\begin{figure}[!htbp]
\centering
    \includegraphics[width=0.43\textwidth]{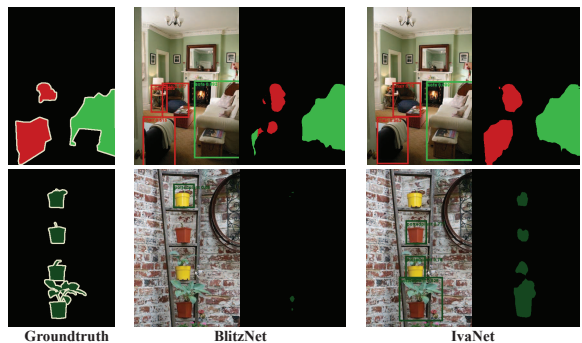}
    \vspace{-0.3cm}
   \caption{Visual results of BlitzNet and the presented IvaNet. The left and the right is detections and segmentations in each paired of images, respectively.}
\label{fig:visual_results}
\vspace{-0.3cm}
\end{figure}

From the visual results in Fig.~\ref{fig:visual_results}, we can observe that the BlitzNet is ineffective to segment small objects, we argue the reason for such failures is that much meaningless information is integrated into lower layers by the full top-down module, while the IvaNet can avoid this case and performs better.

\subsection{Limitation}
\label{sec:limitation}

\begin{figure}[!htbp]
\centering
    \includegraphics[width=0.41\textwidth]{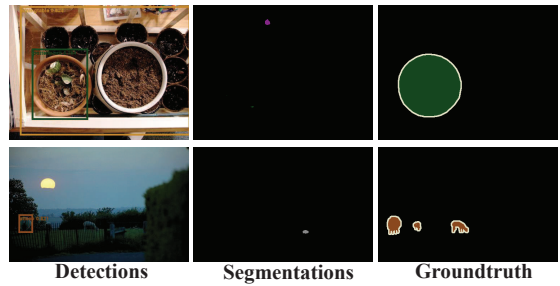}
    \vspace{-0.3cm}
   \caption{Failure cases of the proposed IvaNet.}
\label{fig:failure_case}
\vspace{-0.3cm}
\end{figure}

Despite that the proposed IvaNet has achieved the compelling results in many cases, it still has some limitations.
For instance, IvaNet can detect some objects but fail to segment them as shown in Fig.~\ref{fig:failure_case}.
We argue that the reason behind such failure case is that the limited context is introduced by our model, as semantic segmentation requires more context than object detection.

\section{CONCLUSION}
\label{sec:conclusion}
We present a multi-task framework IvaNet for effectively solving object detection and semantic segmentation in an efficient way. With the help of LTD modules, IvaNet has achieved superior or comparable results on the PASCAL VOC and MS COCO datasets. Due to the limited context of our IvaNet, we are going to adopting some atrous convolutional layers in the future.

\bibliography{references}

\begin{thebibliography}{10}

\bibitem{dssd}
Cheng-Yang Fu, Wei Liu, Ananth Ranga, Ambrish Tyagi, and Alexander~C Berg,
\newblock ``Dssd: Deconvolutional single shot detector,''
\newblock {\em arXiv preprint arXiv:1701.06659}, 2017.

\bibitem{deconvnet}
Hyeonwoo Noh, Seunghoon Hong, and Bohyung Han,
\newblock ``Learning deconvolution network for semantic segmentation,''
\newblock in {\em Proceedings of the IEEE international conference on computer
  vision}, 2015, pp. 1520--1528.

\bibitem{blitznet}
Nikita Dvornik, Konstantin Shmelkov, Julien Mairal, and Cordelia Schmid,
\newblock ``Blitznet: A real-time deep network for scene understanding,''
\newblock in {\em ICCV 2017-International Conference on Computer Vision}, 2017,
  p.~11.

\bibitem{Local_Top_Down}
Shihua Huang, Lu~Wang, Peiyu Yang, and Qingxu Deng,
\newblock ``A local top-down module for object detection with multi-scale
  features,''
\newblock in {\em Chinese Conference on Pattern Recognition and Computer Vision
  (PRCV)}. Springer, 2018, pp. 65--77.

\bibitem{ssd}
Wei Liu, Dragomir Anguelov, Dumitru Erhan, Christian Szegedy, Scott Reed,
  Cheng-Yang Fu, and Alexander~C Berg,
\newblock ``{SSD}: Single shot multibox detector,''
\newblock in {\em European conference on computer vision}. Springer, 2016, pp.
  21--37.

\bibitem{resnet}
Kaiming He, Xiangyu Zhang, Shaoqing Ren, and Jian Sun,
\newblock ``Deep residual learning for image recognition,''
\newblock in {\em Proceedings of the IEEE conference on computer vision and
  pattern recognition}, 2016, pp. 770--778.

\bibitem{fcn}
Jonathan Long, Evan Shelhamer, and Trevor Darrell,
\newblock ``Fully convolutional networks for semantic segmentation,''
\newblock in {\em Proceedings of the IEEE conference on computer vision and
  pattern recognition}, 2015, pp. 3431--3440.

\bibitem{rcnn}
Ross Girshick, Jeff Donahue, Trevor Darrell, and Jitendra Malik,
\newblock ``Rich feature hierarchies for accurate object detection and semantic
  segmentation,''
\newblock in {\em Proceedings of the IEEE conference on computer vision and
  pattern recognition}, 2014, pp. 580--587.

\bibitem{fast}
Ross Girshick,
\newblock ``Fast r-cnn,''
\newblock in {\em Proceedings of the IEEE international conference on computer
  vision}, 2015, pp. 1440--1448.

\bibitem{faster}
Shaoqing Ren, Kaiming He, Ross Girshick, and Jian Sun,
\newblock ``Faster r-cnn: Towards real-time object detection with region
  proposal networks,''
\newblock in {\em Advances in neural information processing systems}, 2015, pp.
  91--99.

\bibitem{ss}
Jasper~RR Uijlings, Koen~EA Van De~Sande, Theo Gevers, and Arnold~WM Smeulders,
\newblock ``Selective search for object recognition,''
\newblock {\em International journal of computer vision}, vol. 104, no. 2, pp.
  154--171, 2013.

\bibitem{deeplabv2}
Liang-Chieh Chen, George Papandreou, Iasonas Kokkinos, Kevin Murphy, and Alan~L
  Yuille,
\newblock ``Deeplab: Semantic image segmentation with deep convolutional nets,
  atrous convolution, and fully connected crfs,''
\newblock {\em IEEE transactions on pattern analysis and machine intelligence},
  vol. 40, no. 4, pp. 834--848, 2018.

\bibitem{pspnet}
Hengshuang Zhao, Jianping Shi, Xiaojuan Qi, Xiaogang Wang, and Jiaya Jia,
\newblock ``Pyramid scene parsing network,''
\newblock in {\em IEEE Conf. on Computer Vision and Pattern Recognition
  (CVPR)}, 2017, pp. 2881--2890.

\bibitem{ubernet}
Iasonas Kokkinos,
\newblock ``Ubernet: Training a universal convolutional neural network for
  low-, mid-, and high-level vision using diverse datasets and limited
  memory,''
\newblock in {\em Proceedings of the IEEE Conference on Computer Vision and
  Pattern Recognition}, 2017, pp. 6129--6138.

\bibitem{mask}
Kaiming He, Georgia Gkioxari, Piotr Doll{\'a}r, and Ross Girshick,
\newblock ``Mask r-cnn,''
\newblock in {\em Computer Vision (ICCV), 2017 IEEE International Conference
  on}. IEEE, 2017, pp. 2980--2988.

\bibitem{tensorflow}
Mart{\'\i}n Abadi, Paul Barham, Jianmin Chen, Zhifeng Chen, Andy Davis, Jeffrey
  Dean, Matthieu Devin, Sanjay Ghemawat, Geoffrey Irving, Michael Isard,
  et~al.,
\newblock ``Tensorflow: a system for large-scale machine learning.,''
\newblock in {\em OSDI}, 2016, vol.~16, pp. 265--283.

\bibitem{Adam}
Diederik~P Kingma and Jimmy Ba,
\newblock ``Adam: A method for stochastic optimization,''
\newblock {\em arXiv preprint arXiv:1412.6980}, 2014.

\bibitem{extra_ann}
Bharath Hariharan, Pablo Arbel{\'a}ez, Lubomir Bourdev, Subhransu Maji, and
  Jitendra Malik,
\newblock ``Semantic contours from inverse detectors,''
\newblock 2011.

\bibitem{mscoco}
Tsung-Yi Lin, Michael Maire, Serge Belongie, James Hays, Pietro Perona, Deva
  Ramanan, Piotr Doll{\'a}r, and C~Lawrence Zitnick,
\newblock ``Microsoft coco: Common objects in context,''
\newblock in {\em European conference on computer vision}. Springer, 2014, pp.
  740--755.

\bibitem{gcn}
Chao Peng, Xiangyu Zhang, Gang Yu, Guiming Luo, and Jian Sun,
\newblock ``Large kernel matters¡ªimprove semantic segmentation by global
  convolutional network,''
\newblock in {\em Computer Vision and Pattern Recognition (CVPR), 2017 IEEE
  Conference on}. IEEE, 2017, pp. 1743--1751.

\end{thebibliography}
\bibliographystyle{IEEEbib}

\end{document}